\documentclass{article} % For LaTeX2e
\usepackage{nips12submit_e,times}
\usepackage{graphicx}
\usepackage{amsmath, amssymb}
\usepackage{natbib}
\usepackage{bm}
\usepackage{caption}
\usepackage{subcaption}
\usepackage{multirow}
\usepackage{ctable}

\title{Binary Classifier Calibration: A Bayesian Non-Parametric Approach}

\author{
Mahdi Pakdaman Naeini \\
Intelligent System Program\\
University of Piuttsburgh\\
\texttt{pakdaman@cs.pitt.edu} \\
\And
Gregory F. Cooper \\
Department of Biomedical Informatics \\
University of Pittsburgh \\
\texttt{gfc@pitt.edu} \\
\AND
Milos Hauskrecht \\
Computer Science Department \\
University of Pittsburgh \\
\texttt{milos@cs.pitt.edu} \\
}

\nipsfinalcopy % Uncomment for camera-ready version

\begin{document} 
\maketitle 
\section*{Abstract}
A set of probabilistic predictions is well calibrated if the events that are
predicted to occur with probability \textit{$p$} do in fact occur about \textit{$p$} fraction of the time. 
Well calibrated predictions are particularly important when machine learning models are 
used in decision analysis.  This paper presents two new non-parametric methods for calibrating 
outputs of binary classification models:  a method based on the Bayes optimal selection and a method 
based on the Bayesian model averaging. The advantage of these methods is that
they are independent of the algorithm used to learn a predictive model, and they can be applied in a post-processing step, 
after the model is learned.  This makes them applicable to a wide variety of machine learning models 
and methods. These calibration methods, as well as other methods, are tested on
a variety of datasets in terms of both discrimination and calibration
performance. The results show the methods either outperform or are comparable in
performance to the state-of-the-art calibration methods. 
\section{Introduction}

This paper focuses on the development of probabilistic calibration methods for
probabilistic prediction tasks. Traditionally, machine-learning research has
focused on the development of methods and models for improving discrimination,
rather than on methods for improving calibration. However, both are very
important. Well-calibrated predictions are particularly important in decision
making and decision analysis
\cite{niculescu2005predicting,zadrozny2001obtaining,zadrozny2002transforming}.
Miscalibrated models, which overestimate or underestimate the probability of outcomes, may lead to making suboptimal
decisions.

Since calibration is often not a priority, many prediction models learned by
machine learning methods may be miscalibrated. The objective of this work is to
develop general but effective methods that can address the miscalibration
problem. Our aim is to have methods that can be used independent of the
prediction model and that can be applied in the post-processing step after the
model is learned from the data. This approach frees the designer of the machine
learning model from the need to add additional calibration measures and terms
into the objective function used to learn the model.  Moreover, all modeling
methods make assumptions, and some of those assumptions may not hold in a given
application, which could lead to miscalibration.  In addition, limited training
data can negatively affect a model's calibration performance.

Existing calibration methods can be divided into parametric and non-parametric
methods. An example of a parametric method is Platt�s method that applies a
sigmoidal transformation that maps the output of a model (e.g., a posterior
probability) \cite{platt1999probabilistic} to a new probability  that is
intended to be better calibrated. The parameters of the sigmoidal transformation
function are learned  using the maximum likelihood estimation framework. A
limitation of the sigmoidal function is that it  is symmetric and does not work
well for highly biased distributions \cite{jiang2012calibrating}. The most
common non-parametric methods are based either on binning
\cite{zadrozny2001obtaining} or isotonic regression \cite{ayer1955empirical}.
Briefly, the binning approach divides the observed outcome predictions into k
bins; each bin is associated with a new probability value that is derived from
empirical estimates. The isotonic regression algorithm is a special adaptive
binning approach that assures the isotonicity (monotonicity) of the probability estimates.

In this paper we introduce two new Bayesian non-parametric
calibration methods. The first one, the \textit{Selection over Bayesian
Binnings} $(SBB)$, uses dynamic programming to efficiently search over
all possible binnings of the posterior probabilities within a training set in
order to select the Bayes optimal binning according to a scoring measure. The
second method, \textit{Averaging over Bayesian Binnings} $(ABB)$, generalizes
$SBB$ by performing model averaging over all possible binnings. The advantage of
these Bayesian methods over existing calibration methods is that they have more
stable, well-performing behavior under a variety of conditions.

Our probabilistic calibration methods can be applied in two prediction settings.
First, they can be used to convert the outputs of discriminative classification
models, which have no apparent probabilistic interpretation, into posterior
class probabilities. An example is an
SVM that learns a discriminative model, which does not have a direct
probabilistic interpretation. Second, the calibration methods can be applied to
improve the calibration of predictions of a probabilistic model that is
miscalibrated. For example, a Na\"{i}ve Bayes (NB)  model is a probabilistic
model, but its class posteriors are often miscalibrated due to unrealistic independence
assumptions \cite{niculescu2005predicting}. The methods we describe are shown
empirically to improve the calibration of NB models without reducing its
discrimination. The methods can also work well on calibrating models that are
less egregiously miscalibrated than are NB models.

The remainder of this paper is organized as follows. Section \ref{Methods}
describes the methods that we applied to perform post-processing calibration.
Section \ref{ExperimentalMethod} describes the experimental setup that 
we used in evaluating the calibration methods. The results of the
experiments are presented in Section \ref{ExperimentalResults}. 
Section \ref{discussion} discusses the results and describes the advantages and
disadvantages of proposed methods in comparison to other calibration methods. Finally, Section
\ref{conclusion} states conclusions, and describes
several areas for future work.

\section{Methods}  
\label{Methods} 
\interfootnotelinepenalty=10000

In this section we present two new Bayesian non-parametric methods for binary
classifier calibration that generalize the histogram-binning calibration 
method \cite{zadrozny2001obtaining} by considering all possible binnings of the
training data. The first proposed method, which is a hard binning classifier
calibration method, is called \textit{Selection over Bayesian}  Binnings $(SBB)$. 
We also introduce a new soft binning method that generalizes $SBB$ by model averaging over all possible binnings; 
it is called \textit{Averaging over Bayesian} Binnings $(ABB)$. There are two main challenges here. 
One is how to score a binning. We use a Bayesian score. 
The other is how to efficiently search over such a large space of binnings. We
use dynamic programming to address this issue.

\subsection{Bayesian Calibration Score} 

Let $p_{in}^i$ and $Z_i$ define respectively an uncalibrated classifier prediction and 
the true class of the $i$'th instance . Also, let $D$ define the set of all training 
instances $(p_{in}^i , Z _i)$. In addition, let $S$ be the \emph{sorted} set of all 
uncalibrated classifier predictions $\{p_{in}^1,p_{in}^2,\ldots,p_{in}^N\}$ and
$S_{l,u}$ be a list of the first elements of $S$, starting at $l$'th index and
ending at $u$'th index, and let $Pa$ denote a binning of $S$. A binning
model $M$ induced by the training set is defined as:
\begin{equation}
	M \equiv \{B, S, Pa, \Theta\},
\end{equation}
where, $B$ is the number of bins over the set $S$ and $\Theta$ is the set of all
the calibration model parameters $\Theta = \{ \theta_1,\ldots, \theta_B\}$, 
which are defined as follows. For a bin $b$, which is determined by $S_{l_b,u_b}$, the distribution of the
class variable $P(Z=1|B = b)$ is modeled as a binomial distribution with parameter $\theta_b$.
Thus, $\Theta$ specifies all the binomial distributions for all the existing bins in $Pa$. 
We note that our binning model is motivated by the model introduced in
\cite{jonathan12application} for variable discretization, which is
here customized to perform classifier calibration. We score a binning model $M$ as follows:
 
\begin{equation}
\label{Eq-Score} 
Score(M) = P(M) \cdot P(D|M) 
\end{equation}

The marginal likelihood $P(D|M)$ in Equation \ref{Eq-Score} is derived using the
marginalization of the joint probability of $P(D,\Theta)$ over all parameter 
space according to the following equation: 

\begin{equation}
\label{Eq-MarginalLikelihood}
P(D|M) = \int_{\Theta} P(D|M,\Theta)P(\Theta|M)d_\Theta
\end{equation}

% Equation \ref{Eq-MarginalLikelihood} has the following closed-form solution, under 
% reasonable assumptions that are described in \cite{heckerman1995learning}:

Equation \ref{Eq-MarginalLikelihood} has a closed form solution under the
following assumptions: (1) All samples are i.i.d and the class distribution
$P(Z|B=b )$ ,which is class distribution for instances locatd in bin number
$b$, is modeled using a binomial distribution with parameter $\theta_b$, (2) the
distribution of class variables over two different bins are independent of each
other, and (3) the prior distribution over binning model parameters $\theta$s
are modeled using a $Beta$ distribution. We also assume that the parameters of
the $Beta$ distribution $\alpha$ and $\beta$ are both equal to one, which means
corresponds to a uniform distribution over each $\theta_b$. The closed form
solution to the marginal likelihood given the above assumptions is as follows
\cite{heckerman1995learning}:

\begin{equation} 
\label{Eq-MLcloseform}
 P(D|M) = \prod_{b=1}^{B}\frac{n_{b0}!\hspace*{0.2em} n_{b1}!}{(n_b+1)!},
\end{equation}

where $n_b$ is the total number of training instances located in bin $b$.
Also, $n_{b0}$ and $n_{b1}$ are respectively the number of class \textit{zero}
and class \textit{one} instances among all $n_b$ training instances in 
bin $b$.

The term $P(M)$ in Equation \ref{Eq-Score} specifies the prior probability of a
binning of calibration model $M$. It can be interpreted as a structure prior,
which we define as follows. Let $Prior(k)$ be the prior probability of there
being a partitioning boundary between $p_{in}^k$ and $p_{in}^{k+1}$ in the binning given by model $M$, and 
model it using a $Poisson$ distribution with parameter $\lambda$.

Consider the prior probability for the presence of bin $b$, which contains the
sequence of training instances $S_{l_b,u_b}$  according to model $M$. 
Assuming independence of the appearance of partitioning boundaries, we can calculate 
the prior of the boundaries defining bin $b$ by using the $Prior$ function as follows:

\begin{equation}
\label{Eq-Prior}
Prior(u_b)\left(\prod_{k=l_b}^{u_b-1}(1-Prior(k)) \right)
\end{equation}
where the product is over all training instances from $S_{l_b}$  to $S_{u_b - 1}$ ,  inclusive.

Combining Equations \ref{Eq-Prior} and \ref{Eq-MLcloseform} into Equation
\ref{Eq-Score}, we obtain the following Bayesian score for calibration model $M$:

\begin{equation}
\label{Eq-BayesianScore}
\scriptsize{
Score(M) = \prod_{b=1}^{B}\left[
Prior(u_b)\left(\prod_{k=l_b}^{u_b-1}(1-Prior(k)) \right)
\frac{n_{b0}!\hspace*{0.2em} n_{b1}!}{(n_b+1)!}  \right]}
\end{equation}

\subsection{The $SBB$ and $ABB$ models}

We can use the above Bayesian score to perform model selection or model averaging. 
Selection involves choosing the best partitioning model $M_{opt}$ and 
calibrating a prediction $x$ as $P(x)= P(x|M_{opt})$. As mentioned, we call this 
approach \textit{Selection over Bayesian Binnings} $(SBB)$. Model averaging
involves calibrating predictions over all possible binnings. We call this 
approach \textit{Averaging over Bayesian Binnings} $(ABB)$ model. A calibrated
prediction in $ABB$ is derived as follows:

\begin{equation}
\label{Eq-BayesianAveraging}
\begin{split}
	P(x)& = \sum_{i=1}^{2^{N-1}}P(M_i|D)P(x|M_i) \\
	    &   \propto \sum_{i=1}^{2^{N-1}}Score(M_i)P(x|M_i),
\end{split}
\end{equation}

where $N$ is the total number of predictions in $D$ (i.e., training instances). 

Both $(SBB)$ and $(ABB)$ consider all possible binnings of the $N$ predictions
in $D$, which is exponential in $N$. Thus, in general, a brute-force approach is not computationally tractable. 
Therefore, we apply dynamic programming, as described in the next two sections.

\subsection{Dynamic Programming Search of $SBB$}

This section summarizes the dynamic programming method used in $SBB$. 
It is based on the dynamic-programming-based discretization method 
described in \cite{jonathan12application}. Recall that $S$ is the  \emph{sorted}
set of all uncalibrated classifier's outputs $\{p_{in}^1,p_{in}^2,\ldots,p_{in}^N\}$  in the training data set. 
Let $S_{1,u }$ define the prefix of set $S$ including the set of the first $u$ uncalibrated 
estimates $\{p_{in}^1,p_{in}^2,\ldots,p_{in}^u\}$. Consider finding the optimal binning 
models $M_{1,u}$ corresponding to the subsequence $S_{1,u}$ for $u \in {1,2,\ldots,N}$ of the set $S$. 
Assume we have already found the highest score binning of these models $M_{1,1}, M_{1,2},\ldots,M_{1,u-1}$, 
corresponding to each of the subsequences $S_{1,1}, S_{1,2},\ldots,S_{1,u-1}$. 
Let $V^f_1,V^f_2,\ldots,V^f_{u-1}$  denote the respective scores of the optimal binnings of these models. 
Let $Score_{i,u}$ be the score of subsequence $\{p_{in}^i,p_{in}^2,\ldots,p_{in}^u\}$ when it is 
considered as a single bin in the calibration model $M_{1,u}$. For all $l$ from $u$ to $1$, $SBB$ 
computes $V^f_{l-1}\times Score_{l,u}$, which is the score for the highest scoring 
binning $M_{1,u}$ of set $S_{1,u}$ for which 
subsquence $S_{l,u}$  is considered as a single bin. Since this binning score is derived 
from two other scores , we call it a \textit{composite score} of the binning
model $M_{1,u}$. The fact that this composite score is a product of two scores
follows from the decomposition of Bayesian scoring measure we are using, as
given by Equation \ref{Eq-BayesianScore}. In particular, both the prior and
marginal likelihood terms on the score are decomposable.

In finding the best binning model $M_{1,u}$, $SBB$ chooses the maximum composite
score over all $l$, which corresponds to the optimal binning for the training data subset $S_{1,u}$; 
this score is stored in $V^f_u$. By repeating this process from $1$ to $N$, $SBB$ derives the 
optimal binning of set $S_{1,N}$, which is the best binning over all possible binnings. 
As shown in \cite{jonathan12application}, the computational time complexity of the above dynamic
programming procedure is $O(N^2)$.

\subsection{Dynamic Programming Search of $ABB$}  

The dynamic programming approach used in $ABB$ is based on the above dynamic
programming approach in $SBB$. It focuses on calibrating a particular instance $P(x)$. 
Thus, it is an instance-specific method. The $ABB$ algorithm uses the $decomposability$ property 
of the Bayesian binning score in Equation \ref{Eq-BayesianScore}. Assume we have already 
found in one forward run (from lowest to highest prediction) of the $SBB$ method the highest 
score binning of the models $M_{1,1}, M_{1,2},\ldots,M_{1,N}$, which correspond to each of 
the subsequences $S_{1,1}, S_{1,2},\ldots,S_{1,N}$, respectively; let the 
values $V^f_1,V^f_2,\ldots,V^f_{N}$ denote the respective scores of the optimal binning for 
these models, which we cache. We perform an analogous dynamic programming procedure in $SBB$ in a 
backward manner (from highest to lowest prediction) and compute the highest score binning of 
these models $M_{N,N}, M_{N-1,N},\ldots,M_{1,N}$, which correspond to each of the 
subsequences $S_{N,N}, S_{N-1,N},\ldots,S_{1,N}$, respectively; let the 
values $V^b_N,V^b_{N-1},\ldots,V^b_{1}$  denote the respective scores of the optimal binning 
for these models, which also cache. Using the decomposability property of the binning 
score given by \ref{Eq-BayesianScore}, we can write the Bayesian model averaging estimate given 
by Equation \ref{Eq-BayesianAveraging} as follows:

\begin{equation}
P(x) \propto \sum_{1 \leq l \leq u \leq N} \left(V^f_{l-1} \times Score_{l,u} \times V^b_{u+1} \times 
\hat{p}_{l,u}(x) \right ) 
\end{equation} 

where $\hat{p}_{l,u}(x)$ is obtained using the frequency\footnote{we actually
use smoothing of these counts, which is consistent with the Bayesian priors in the scoring function}
of the training instances in the bin containing the predictions $S_{l,u}$. 
Remarkably, the dynamic programming implementation of $ABB$ is also $O(N^2)$. 
However, since it is instance specific, this time complexity holds for 
each prediction that is to be calibrated (e.g., each prediction in a test set).
To address this problem, we can partition the interval [0, 1] into $R$ equally 
spaced bins and stored the ABB output for each of those bins. The training time
is therefore $O(R N^2)$. During testing, a given $p_{in}$ is mapped to one of
the $R$ bins and the stored calibrated probability is retrieved, which can all be
done in $O(1)$ time.

\section{Experimental Setup}
\label{ExperimentalMethod}

\begin{figure}[h] 
\begin{minipage}{16pc}
\centering 
  		\includegraphics[scale= 0.29]{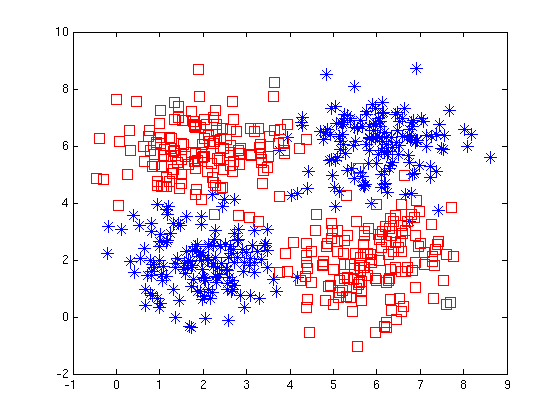}
  		\caption{XOR Configuration}
        \label{fig:Nonlinear1} 
\end{minipage}\hspace{2pc}%
\begin{minipage}{16pc}
   	\centering
  		\includegraphics[scale=0.29]{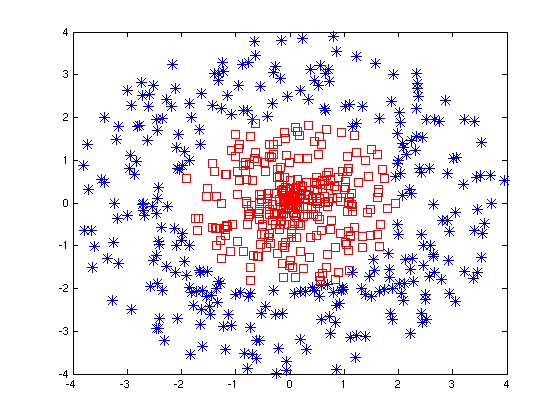}
  		\caption{Circular Configuration }
        \label{fig:NonLinear2}

\end{minipage}
\label{fig:SimulationData}
\end{figure}

This section describes the set of experiments that we performed to evaluate the calibration 
methods described above. To evaluate the calibration performance of each method, 
we ran experiments using both simulated data and real data. In our experiments
on simulated data, we used logistic regression (LR) as the base classifier, whose
predictions are to be calibrated.
The choice of logistic regression was made to let us compare our results with the state-of-the-art method $ACP$, 
which as published is tailored for LR. For the simulated data, we used one dataset 
in which the outcomes were linearly separable and two other datasets in
which they were not. Also, in the simulation data we used  $600$ randomly
generated instances for training LR model, $600$ random instances for learning
calibration-models, and $600$ random instances for testing the models 
\footnote{Based on our experiments the separation between training set and calibration
set is not necessary. However, \cite{zadrozny2001obtaining} state that for the histogram model it
is better to use another set of instances for calibrating 
the output of classifier; thus, we do so here.} The scatter plots of the
two linearly non-separable simulated datasets are shown in Figures 
[\ref{fig:Nonlinear1}, \ref{fig:NonLinear2} ].

% \begin{figure}
%   \centering
%   	\begin{subfigure} [b]{0.4\textwidth}
%   		\centering
%   		\includegraphics[scale=0.29]{./Nonlinear_Scatter_XOR.png}
%   		\caption{XOR Configuration}
%         \label{fig:Nonlinear1}
% 	\end{subfigure}
%     \begin{subfigure} [b]{0.4\textwidth}
%     	\centering
%   		\includegraphics[scale=0.29]{./Nonlinear_Scatter_Circular.png}
%   		\caption{Circular Configuration}
%         \label{fig:NonLinear2}
% 	\end{subfigure}
%   \caption{Scatter plots of the simulated data}
%   \label{fig:SimulationData}
% \end{figure}

We also performed experiments on three different sets of real binary
classification data. The first set is the UCI Adult dataset. The prediction task
is a binary classification problem to predict whether a person makes over
\textdollar50K a year using his or her demographic information. From the
original Adult dataset, which includes $48842$ total instances with $14$
real and categorical features, after removing the instances with missing
values, we used $2000$ instances for training classifiers, $600$ for
calibration-model learning, and $600$ instances for testing. 

We also used the UCI SPECT dataset, which is a small biomedical binary
classification dataset. SPECT allows  us to examine how well each calibration method performs when the
calibration dataset is small in a real application.
The dataset involves the diagnosis of cardiac Single Proton Emission Computed Tomography (SPECT) images.
Each of the patients is classified into two categories: normal or abnormal.
This dataset consists of $80$ training instances, with an equal number of positive and 
negative instances, and $187$ test instances with only $15$ positive instances.
The SPECT dataset includes $22$ binary features. Due to the small 
number of instances, we used the original training data 
as both our training and calibration datasets, and we  used the
original test data as our test dataset.

For the experiments on the Adult and SPECT datasets, we used three different
classifiers: LR, na\"{i}ve Bayes, and SVM with polynomial kernels. The choice of
the LR model allows us to include the ACP method in the comparison, because as
mentioned it is tailored to LR. Na\"{i}ve Bayes is a well-known,
simple, and practical classifier that often achieves good discrimination
performance, although it is usually not well calibrated. We included SVM because
it is a relatively modern classifier that is being frequently applied.

The other real dataset that we used for evaluation contains clinical findings
(e.g., symptoms, signs, laboratory results) and outcomes for patients with
community acquired pneumonia (CAP)
\cite{fine1997prediction}.
The classification task we examined involves using patient findings to predict dire patient outcomes,
such as mortality or serious medical complications. The CAP dataset includes a
total of $2287$ patient cases (instances) that we divided into $1087$ instances for training of
classifiers, $600$ instances for learning calibration models, and $600$
instances for testing the calibration models. The data includes $172$ discrete
and $43$ continuous features. For our experiments on the na\"{i}ve Bayes model,
we just used the discrete features of data, and for the experiments on SVM we used all
$215$ discrete and continuous features. Also, for applying LR model to this
dataset, we first used PCA feature transformation because of the high
dimensionality of data and the existing correlations among some features, which
produced unstable results due to singularity issues.

\section{Experimental Results}
\label{ExperimentalResults}
\vspace{-2 mm}

This section presents experimental results of the calibration 
methods when applied to the datasets described in the previous section. We show the 
performance of the methods in terms of both calibration and discrimination, 
since in general both are important. Due to a lack of space, we do not include here the results for
the linearly separable data; however, we note that the results for each of the calibration 
methods and the base classifier was uniformly excellent across the different evaluation 
measures that are described below.

For the evaluation of the calibration methods, we used $5$ different measures. 
The first two measures are Accuracy (Acc) and the Area Under the ROC Curve
(AUC), which measure discrimination. The three other measures are the Root Mean
Square Error (RMSE), Expected Calibration Error (ECE), and Maximum Calibration Error (MCE).
These measures evaluate calibration performance. The $ECE$
and $MCE$ are simple statistics that measure calibration relative to the ideal
reliability diagram \cite{degroot1983comparison,niculescu2005predicting}.
In computing these measures, the predictions are partitioned to ten bins
 $\{[0,.1),[.1,.2),\ldots,[.9,1]\}$.
The predicted value of each test instance falls into one of the bins.  
The $ECE$ calculates Expected Calibration Error over the bins, and $MCE$
calculates the Maximum Calibration Error among the bins, using empirical
estimates as follows:
\begin{eqnarray*}
ECE &=&  \sum_{i=1}^{10} P(i) \cdot \left|o_i-e_i \right| \\
MCE &=&  \max \left( \left| o_i-e_i \right| \right)
\end{eqnarray*}  
where $o_i$ is the true fraction of positive instances in bin $i$, $e_i$ is the mean of the
post-calibrated probabilities for the instances in bin $i$, and $P(i)$  is
the empirical probability (fraction) of all instances that fall into bin $i$. The
lower the values of $ECE$ and $MCE$, the better is the calibration of a model.

The Tables $[\ref{tab:SimulatedData1}, \ref{tab:SimulatedData2}, \ldots, \ref{tab:Pnuemonia_SVM}]$ show 
the comparisons of different methods with respect to evaluation measures on the simulated and real datasets. 
In these tables in each row we show in bold the two methods that achieved the best performance with respect 
to a specified measure.  

As can be seen, there is no superior method that outperforms all the others in
all data sets on all measures. However, SBB and ABB are superior to Platt and isotonic
regression in all the simulation datasets. We discuss the reason why 
in Section \ref{discussion}. Also, SBB and ABB perform as well 
or better than isotonic regression and the Platt method on the real data sets.

In all of the experiments, both on simulated datasets and real data sets, 
both SBB and ABB generally retain or improve the
discrimination performance of the base classifier, as measured by 
Acc and AUC. In addition, they often
improve the calibration performance of the base classifier in terms of
the $RMSE$, $ECE$ and $MCE$ measures.
\section{Discussion }
\label{discussion}
Having a well-calibrated classifier can be important in practical machine
learning problems. There are different calibration methods in the literature
and each one has its own pros and cons. The Platt method uses a sigmoid as a mapping function. 
The main advantage of Platt scaling method is its fast recall time. However,
the shape of sigmoid function can be restrictive, and it often cannot produce
well calibrated probabilities when the instances are distributed in feature 
space in a biased fashion (e.g. at the extremes, or all near separating hyper
plane) \cite{jiang2012calibrating}.
  
Histogram binning is a non-parametric method which makes no
special assumptions about the shape of mapping function. However, it has several
limitations, including the need to define the number of bins and the fact 
that the bins remain fixed over all predictions \cite{zadrozny2002transforming}.

Isotonic regression-based calibration is another non-parametric calibration
method, which requires that the mapping 
(from pre-calibrated predictions to post-calibrated ones) is chosen from
the class of all isotonic (i.e., monotonicity increasing) functions
\cite{niculescu2005predicting,zadrozny2002transforming}. Thus, it is less
restrictive than the Platt calibration method. The \textit{pair adjacent
violators} (PAV) algorithm is one instance of an isotonic regression algorithm
\cite{ayer1955empirical}. The PAV algorithm can be considered as 
a binning algorithm in which the boundaries of the bins are
chosen according to how well the classifier ranks the
examples\cite{zadrozny2002transforming}.  It has been shown that Isotonic
regression performs very well in comparison to other calibration methods in real
datasets \cite{niculescu2005predicting, caruana2006empirical,
zadrozny2002transforming}. Isotonic regression has some limitations, however.
The most significant limitation of the isotonic regression is its isotonicity (monotonicity) assumption.
As seen in Tables [\ref{tab:SimulatedData1}, \ref{tab:SimulatedData2}] in the 
simulation data, when the isotonicity assumption
is violated through the choice of classifier and the nonlinearity of data, 
isotonic regression performs relatively poorly, in terms of improving the discrimination and
calibration capability of a base classifier.
The violation of this assumption can happen in real data secondary to 
the choice of learning  models and algorithms.

A classifier calibration method called \textit{adaptive calibration
of predictions} (ACP) was recently introduced \cite{jiang2012calibrating}. 
A given application of ACP is tied to a particular model $M$, such as a logistic regression model,
that predicts a binary outcome $Z$.
ACP requires a $95\%$ confidence interval (CI) around a particular prediction
$p_{in}$ of $M$. ACP adjusts the CI and uses it to define a bin.
It sets $p_{out}$  to be the fraction of positive outcomes $(Z=1)$ 
among all the predictions that fall within the bin.
On both real and synthetic datasets, ACP achieved better calibration 
performance than a variety of other calibration methods,
including simple histogram binning, Platt scaling, 
and isotonic regression \cite{jiang2012calibrating}.
The ACP post-calibration probabilities also achieved among 
the best levels of discrimination, 
according to the AUC. ACP has several limitations, however. 
First, it requires not only probabilistic predictions, but also a 
statistical confidence interval ($CI$) around 
each of those predictions, which makes it tailored to specific classifiers, such as logistic
regression \cite{jiang2012calibrating}. Second, based on a $CI$ around a given prediction $p_{in}$, 
it commits to a single binning of the data around that prediction;
it does not consider alternative binnings that might yield a better calibrated $p_{out}$. 
Third, the bin it selects is symmetric around $p_{in}$ by construction, which may not optimize calibration.
Finally, it does not use all of the training data, but rather only uses those 
predictions within the confidence interval around $p_{in}$. As one can see from the tables, 
ACP performed well when logistic regression is the base classifier, both in simulated and real datasets. 
SBB and ABB performed as well or better than ACP in both simulation and real data sets. 

In general, the SBB and ABB algorithms appear promising, especially ABB, which overall outperformed SBB. 
Neither algorithm makes restrictive (and potentially unrealistic) assumptions, as does Platt scaling 
and isotonic regression. They also are not restricted in the type of 
classifier with which they can apply, unlike ACP.  

The main disadvantage of SBB and ABB is their running time. If $N$ is the number of training instances, 
then SBB has a training time of $O(N^2)$, due to its dynamic programming 
algorithm that searches over every possible binning, whereas the time complexity
of ACP and histogram binning is $O(N logN)$, and it is $O(N)$ for isotonic regression. 
Also, the cached version of ABB has a
training time of $O(R N^2)$, where $R$ reflects the number of bins being used.   
Nonetheless, it remains practical to use these algorithms to perform calibration on a desktop computer 
when using training datasets that contain thousands of instances. In addition, the testing time  
is only $O(b)$ for SBB where $b$ is the number of binnings found by the algorithm
and $O(1)$ for the cached version of ABB. Table \ref{tbl:RunningTime} shows
the time complexity of different methods in learning for N training
instances and recall for only one instance.

\begin{small}
\begin{table}
	\centering
	\caption{Time complexity of calibration methods in learning and recall}
	\scalebox{0.76}{
	\begin{tabular}{p{0.12\textwidth}lllllll}
	      &  \textbf{Platt} &  \textbf{Hist}  &  \textbf{IsoReg} &  \textbf{ACP} &
	       \textbf{SBB}   &  \textbf{ABB}
	      \\
	    \hline
% 	Time      Complexity  & $O(N T + M)$ & $O(N logN + M b)$ & $O(N + M b)$ & $O(N
% 	logN + M b)$ & $O(N^2+M b)$ & $O(R N^2+M)$ \\
	Time   & 
	\multirow{2}{*}{$O(N T) / O(1)$} &
	\multirow{2}{*}{$O(N logN) / O(b)$} &
	\multirow{2}{*}{$O(N) / O(b)$} &
	\multirow{2}{*}{$O(N logN) / O(N)$} &
	\multirow{2}{*}{$O(N^2) / O(b)$} &
	\multirow{2}{*}{$O(R N^2) / O(1)$}\\
	Complexity &  &  &  &  &  & \\ 
	(Learning/Recall) &  &  &  &  &  & \\ 
    %\multicolumn{2}{l}{(Learning/Recall)} &  &  &  &  &  & \\ 
	\hline
	\multicolumn{7}{p{1.2\textwidth}}{\footnotesize Note that N and b are the
	% size
	of training sets and the number of bins found by the method respectively. T	is
	the number of iteration required for convergence in Platt method and R reflects
	the number of bins being used by cached ABB. }
	%\hline
% 	N and M stands for size trainig and test data, and T is the number of iteration
% 	required for convergence. 
	\end{tabular}
	}
	\label{tbl:RunningTime}
\end{table}
\end{small}
\section{Conclusion}
\label{conclusion}
\vspace{-2 mm}

In this paper we introduced two new Bayesian, non-parametric methods for calibrating binary classifiers, 
which are called $SBB$ and $ABB$. Experimental results on simulated and real data support that these methods 
perform as well or better than the other calibration methods that we evaluated.  While the new methods have a 
greater time complexity  than the other calibration methods evaluated here, they nonetheless are efficient enough 
to be applied to training datasets with thousands of instances.  Thus, we believe these new methods are promising 
for use in machine learning, particularly when calibrated probabilities are important, such as in decision analyses.

In future work, we plan to explore how the two new methods perform when using Bayesian model averaging
over the hyper parameter $\lambda$. We also will extend them to perform multi-class calibration.
Finally, we plan to investigate the use of calibration methods on posterior probabilities that
are inferred from models that represent joint probability distributions,
such as maximum-margin Markov-network models
\cite{roller2004max,zhu2008laplace,zhu2009medlda}.
%\dbltextfloatsep 1.5pt
%\dblfloatsep 3.5pt
\belowcaptionskip 8pt
\abovecaptionskip 3pt
%\intextsep 1pt

\begin{small}
\begin{table}[!ht]
	\caption{Experimental Results on Simulated and Real datasets}
    \begin{subtable}[b]{0.5\textwidth}
    \centering
 		  \tabcolsep 4.2pt
		  \caption{Non-Linear XOR configuration results}
		  \resizebox{6.5cm}{!} {
% 		  \scalebox{0.70}{ 
			\begin{tabular}{l|cccccccccccccc}
			%\hline
				  & LR    & ACP   & IsoReg & Platt & Hist  & SBB   & ABB  \\
			\hline
			AUC   & 0.497 & \textbf{0.950} & 0.704 & 0.497 & 0.931 & 0.914 & \textbf{0.941} \\
			%\hdashline[1pt/2pt]
			Acc   & 0.510 & \textbf{0.887} & 0.690 & 0.510 & 0.855 & \textbf{0.887} & \textbf{0.888} \\
			%\hdashline[1pt/2pt]
			RMSE  & 0.500 & \textbf{0.286} & 0.447 & 0.500 & 0.307 & 0.307 & \textbf{0.295} \\
			%\hdashline[1pt/2pt]
			MCE   & 0.521 & \textbf{0.090} & 0.642 & 0.521 & 0.152 & 0.268 & \textbf{0.083} \\
			%\hdashline[1pt/2pt]
			ECE   & 0.190 & \textbf{0.056} & 0.173 & 0.190 & 0.072 & 0.104 & \textbf{0.062} \\
			%\hline
			\end{tabular}
			 }
			\label{tab:SimulatedData1}
	 \end{subtable}
     \begin{subtable}[b]{0.5\textwidth} 
     \centering
 		  \tabcolsep 4.2pt
		  \caption{Non-Linear Circular configuration results}
		  \resizebox{6.5cm}{!} {
% 		  \scalebox{0.7}{ 
			\begin{tabular}{l|cccccccccccccc}
			%\hline
				 & LR    & ACP   & IsoReg & Platt & Hist  & SBB   & ABB \\
			\hline
			AUC   & 0.489 & \textbf{0.852} & 0.635 & 0.489 & 0.827 & 0.816 & \textbf{0.838} \\
			%\hdashline[1pt/2pt]
			Acc   & 0.500 & 0.780 & 0.655 & 0.500 & \textbf{0.795} & \textbf{0.790} & 0.773 \\
			%\hdashline[1pt/2pt]
			RMSE  & 0.501 & \textbf{0.387} & 0.459 & 0.501 & 0.394 & 0.393 & \textbf{0.390} \\
			%\hdashline[1pt/2pt]
			MCE   & 0.540 & 0.172 & 0.608 & 0.539 & \textbf{0.121} & 0.790 & \textbf{0.146} \\
			%\hdashline[1pt/2pt]
			ECE   & 0.171 & 0.098 & 0.186 & 0.171 & \textbf{0.074} & 0.138 & \textbf{0.091} \\
			%\hline
			\end{tabular}%
		     }
		     \label{tab:SimulatedData2}
	 \end{subtable}
%\end{table}
%\end{small}

%\begin{small}
%\begin{table}[!ht]
    \begin{subtable}[b]{0.5\textwidth}
	  \centering
	  \tabcolsep 4.2pt
	  \caption{Adult Na\"{i}ve Bayes }
		\resizebox{6.5cm}{!} {
% 	  \scalebox{0.7}{ 
		\begin{tabular}{l|cccccccccccccc}
		%\hline
		 & NB & IsoReg & Platt & Hist & SBB	 & ABB\\
		\hline
		AUC   & \textbf{0.879} & 0.876 & \textbf{0.879} & 0.877 & 0.849 & \textbf{0.879} \\
		%\hdashline[1pt/2pt]
		Acc   & 0.803 & 0.822 & \textbf{0.840} & 0.818 & \textbf{0.838} & 0.835 \\
		%\hdashline[1pt/2pt]
		RMSE  & 0.352 & \textbf{0.343} & \textbf{0.343} & \textbf{0.341} & 0.345 & \textbf{0.343} \\
		%\hdashline[1pt/2pt]
		MCE   & 0.223 & 0.302 & \textbf{0.092} & 0.236 & 0.373 & \textbf{0.136} \\
		%\hdashline[1pt/2pt]
		ECE   & 0.081 & 0.075 & \textbf{0.071} & 0.078 & 0.114 & \textbf{0.062} \\
		%\hline
		\end{tabular}%
	    }
	  \label{tab:Adult_NB}%
  \end{subtable}
  \begin{subtable}[b]{0.5\textwidth}
		   \centering  
		  \tabcolsep 4.2pt
		  \caption{Adult Linear SVM}
		  \resizebox{6.5cm}{!} {
% 		  \scalebox{0.7}{ 
			\begin{tabular}{l|cccccccccccccc}
			%\hline
			 & SVM  & IsoReg & Platt & Hist  & SBB   & ABB  \\
			\hline
			AUC   & \textbf{0.864} & 0.856 & \textbf{0.864} & \textbf{0.864} & 0.821 & \textbf{0.864} \\
			%\hdashline[1pt/2pt]
			Acc   & 0.248 & \textbf{0.805} & 0.748 & \textbf{0.815} & 0.803 & \textbf{0.805} \\
			%\hdashline[1pt/2pt]
			RMSE  & 0.587 & 0.360 & 0.434 & \textbf{0.355} & 0.362 & \textbf{0.357} \\
			%\hdashline[1pt/2pt]
			MCE   & 0.644 & 0.194 & 0.506 & \textbf{0.144} & 0.396 & \textbf{0.110} \\
			%\hdashline[1pt/2pt]
			ECE   & 0.205 & 0.085 & 0.150 & \textbf{0.077} & 0.108 & \textbf{0.061} \\
			%\hline
			\end{tabular}%
		    }
		  \label{tab:Adult_SVM}%
  \end{subtable}
  \begin{subtable}[b]{0.5\textwidth}
		  \centering
		  \tabcolsep 4.2pt
		  \caption{Adult Logistic Regression}
		  \resizebox{6.5cm}{!} {
% 		  \scalebox{0.6}{
			\begin{tabular}{l|ccccccccccccccc}
			%\hline
			      & LR    & ACP   & IsoReg & Platt & Hist  & SBB   & ABB  \\
			\hline
			AUC   & 0.730 & 0.727 & \textbf{0.732} & 0.730 & 0.743 & 0.699 & 0.731 \\
			%\hdashline[1pt/2pt]
			Acc   & 0.755 & \textbf{0.783} & 0.753 & 0.755 & 0.753 & 0.762 & \textbf{0.762} \\
			%\hdashline[1pt/2pt]
			RMSE  & 0.403 & 0.402 & 0.403 & 0.405 & 0.400 & \textbf{0.401} & \textbf{0.401} \\
			%\hdashline[1pt/2pt]
			MCE   & \textbf{0.126} & 0.182 & 0.491 & 0.127 & 0.274 & 0.649 & \textbf{0.126} \\
			%\hdashline[1pt/2pt]
			ECE   & \textbf{0.075} & \textbf{0.071} & 0.118 & 0.079 & 0.092 & 0.169 & 0.076 \\
			%\hline
			\end{tabular}%
		    }
		  \label{tab:Adult_LR}%
  \end{subtable}
	\begin{subtable}[b]{0.5\textwidth}
		  \centering
		  \tabcolsep 4.2pt
		  \caption{SPECT Na\"{i}ve Bayes}
		  \resizebox{6.5cm}{!} {
% 		  \scalebox{0.7}{ 
			\begin{tabular}{l|cccccccccccccc}
			%\hline
			      & NB    & IsoReg & Platt & Hist  & SBB   & ABB\\
			\hline
			AUC   & \textbf{0.836} & 0.815 & \textbf{0.836} & 0.832 & 0.733 & 0.835 \\
			%\hdashline[1pt/2pt]
			Acc   & 0.759 & \textbf{0.845} & 0.770 & 0.824 & \textbf{0.845} & \textbf{0.845} \\
			%\hdashline[1pt/2pt]
			RMSE  & 0.435 & \textbf{0.366} & 0.378 & 0.379 & \textbf{0.368} & 0.374 \\
			%\hdashline[1pt/2pt]
			MCE   & 0.719 & 0.608 & 0.563 & 0.712 & \textbf{0.347} & \textbf{0.557} \\
			%\hdashline[1pt/2pt]
			ECE   & 0.150 & \textbf{0.141} & 0.148 & \textbf{0.145} & 0.149 & 0.157 \\
			%\hline
			\end{tabular}%
		    }
		  \label{tab:SPECT_NB}%
  \end{subtable}
  \begin{subtable}[b]{0.5\textwidth}
 		   \centering
		  \tabcolsep 4.2pt
		  \caption{SPECT SVM Quadratic kernel}
		  \resizebox{6.5cm}{!} {
% 		  \scalebox{0.7}{ 
			\begin{tabular}{l|cccccccccccccc}
			%\hline
			 & SVM    & IsoReg & Platt & Hist  & SBB   & ABB\\
			\hline
			AUC   & \textbf{0.816} & 0.786 & \textbf{0.816} & 0.766 & 0.746 & 0.810 \\
			%\hdashline[1pt/2pt]
			Acc   & 0.257 & \textbf{0.834} & 0.684 & \textbf{0.845} & 0.813 & 0.813 \\
			%\hdashline[1pt/2pt]
			RMSE  & 0.617 & 0.442 & 0.460 & 0.463 & \textbf{0.398} & \textbf{0.386} \\
			%\hdashline[1pt/2pt]
			MCE   & \textbf{0.705} & \textbf{0.647} & 0.754 & 0.934 & 0.907 & 0.769 \\
			%\hdashline[1pt/2pt]
			ECE   & 0.235 & 0.148 & 0.162 & 0.180 & \textbf{0.128} & \textbf{0.131} \\
			%\hline
			\end{tabular}%
		    }
		  \label{tab:SPECT_SVM}%
  \end{subtable}
  \begin{subtable}[b]{0.5\textwidth}
		   \centering
		  \tabcolsep 4.2pt
		  \caption{SPECT Logistic Regression}
		  \resizebox{6.5cm}{!} {
% 		  \scalebox{0.7}{ 
			\begin{tabular}{l|ccccccccccccccc}
			%\hline
			      & LR    & ACP   & IsoReg & Platt & Hist  & SBB   & ABB   \\
			\hline
			AUC   & \textbf{0.744} & 0.742 & 0.733 & \textbf{0.744} & 0.738 & 0.733 & 0.741 \\
			%\hdashline[1pt/2pt]
			Acc   & \textbf{0.658} & 0.561 & 0.626 & \textbf{0.668} & 0.620 & 0.620 & 0.626 \\
			%\hdashline[1pt/2pt]
			RMSE  & 0.546 & 0.562 & 0.558 & 0.524 & 0.565 & \textbf{0.507} & \textbf{0.496} \\
			%\hdashline[1pt/2pt]
			MCE   & 0.947 & 1.000 & 1.000 & 0.884 & 0.997 & \textbf{0.813} & \textbf{0.812} \\
			%\hdashline[1pt/2pt]
			ECE   & 0.181 & 0.187 & 0.177 & 0.180 & 0.183 & \textbf{0.171} & \textbf{0.173} \\
			%\hline
			\end{tabular}%
		    }
		  \label{tab:SPECT_LR}%
  \end{subtable}
  
  %\begin{small}
%\begin{table}[!ht]
    \begin{subtable}[b]{0.5\textwidth}
	  \centering
	  \tabcolsep 4.5pt
	  \caption{CAP Na\"{i}ve Bayes }
	  \resizebox{6.5cm}{!} {
% 	  \scalebox{0.7}{ 
		\begin{tabular}{l|cccccccccccccc}
		%\hline
		      & NB    & IsoReg & Platt & Hist  & SBB   & ABB \\
		\hline	
		AUC   & \textbf{0.848} & 0.845 & \textbf{0.848} & 0.831 & 0.775 & 0.838 \\
		%\hdashline[1pt/2pt]
		Acc   & 0.730 & \textbf{0.865} & 0.847 & 0.853 & 0.832 & \textbf{0.865} \\
		%\hdashline[1pt/2pt]
		RMSE  & 0.504 & \textbf{0.292} & 0.324 & 0.307 & 0.315 & \textbf{0.304} \\
		%\hdashline[1pt/2pt]
		MCE   & 0.798 & 0.188 & 0.303 & \textbf{0.087} & 0.150 & \textbf{0.128} \\
		%\hdashline[1pt/2pt]
		ECE   & 0.161 & 0.071 & 0.097 & \textbf{0.056} & \textbf{0.067} & \textbf{0.067} \\
		%\hline
		\end{tabular}%	
	    }
	    \label{tab:Pnuemonia_NB}
     \end{subtable}
     \begin{subtable}[b]{0.5\textwidth}
       \centering
		  
		  \tabcolsep 4.5pt
		  \caption{CAP Linear SVM }
		  \resizebox{6.5cm}{!} {
% 		  \scalebox{0.7}{ 
			\begin{tabular}{l|cccccccccccccc}
			%\hline
			      & SVM   & IsoReg & Platt & Hist  & SBB   & ABB \\
			\hline
			AUC   & \textbf{0.858} & \textbf{0.858} & \textbf{0.858} & 0.847 & 0.813 & \textbf{0.863} \\
			%\hdashline[1pt/2pt]
			Acc   & \textbf{0.907} & 0.900 & 0.882 & 0.887 & 0.902 & \textbf{0.908} \\
			%\hdashline[1pt/2pt]
			RMSE  & 0.329 & \textbf{0.277} & 0.294 & 0.287 & 0.285 & \textbf{0.274} \\
			%\hdashline[1pt/2pt]
			MCE   & 0.273 & \textbf{0.114} & 0.206 & \textbf{0.110} & 0.240 & 0.121 \\
			%\hdashline[1pt/2pt]
			ECE   & 0.132 & 0.058 & 0.093 & \textbf{0.057} & 0.083 & \textbf{0.050} \\
			%\hline
			\end{tabular}% 
		    }
		    \label{tab:Pnuemonia_SVM}
     \end{subtable}
     \begin{subtable}[b]{\textwidth}
       \centering
  		\tabcolsep 4.5pt
		  \caption{CAP Logistic Regression }
		  \resizebox{7.5cm}{!} {
% 		  \scalebox{0.7}{ 
			\begin{tabular}{l|cccccccccccccc}
			\centering
		        & LR    & ACP   & IsoReg & Platt & Hist  & SBB   & ABB \\
			\hline
			AUC   & \textbf{0.920} & 0.910 & 0.917 & \textbf{0.920} & 0.901 & 0.856 &
			\textbf{0.921} \\
			Acc   & 0.925 & 0.932 & \textbf{0.935} & 0.928 & 0.897 & \textbf{0.935} & 0.932 \\
			RMSE  & \textbf{0.240} & \textbf{0.240} & \textbf{0.234} & 0.242 & 0.259 & \textbf{0.240} & \textbf{0.240} \\
			MCE   & 0.199 & \textbf{0.122} & 0.286 & \textbf{0.154} & 0.279 & 0.391 & 0.168 \\
			ECE   & \textbf{0.066} & \textbf{0.062} & 0.078 & 0.082 & 0.079 & 0.103 & 0.069 \\
  			\end{tabular}% 
		}
  		\label{tab:Pnuemonia_SVM}
     \end{subtable}
  
\end{table}
\end{small}

\bibliographystyle{plain}

\end{document}